\documentclass[12pt,letterpaper]{article}
\usepackage[a4paper, total={7in, 10in}]{geometry}

\usepackage{graphicx}
\usepackage{helvet}
\usepackage{authblk}
\usepackage{hyperref}
\usepackage{amsmath} 
\usepackage{amssymb} 
\usepackage{orcidlink} 
\usepackage[super,comma,sort&compress]  
   {natbib}

\usepackage{algorithm}
\usepackage{tcolorbox}
\usepackage{xcolor}
\usepackage{multirow}
\usepackage{tabularx}
\usepackage{booktabs} 
\usepackage{amsmath}
\usepackage{amssymb}
\usepackage{mathtools}
\usepackage{amsthm}
\usepackage{algorithm}
\usepackage{algorithmic}

\makeatletter
\renewcommand{\maketitle}{\bgroup\setlength{\parindent}{0pt}
\begin{flushleft}
  \textbf{\@title}
  
  \@author
\end{flushleft}\egroup}
\makeatother

\title{MC-NEST: Enhancing Mathematical Reasoning in Large Language Models leveraging a Monte Carlo Self-Refine Tree}
\date{}

\author[1,*,\orcidlink{0000-0000-0000-0000}]{Gollam Rabby}
\author[2\orcidlink{0000-0002-3782-8069}]{Farhana Keya}
\author[1,2\orcidlink{0000-0002-0698-2864}]{Sören Auer}

\affil[1]{ L3S Research Center, Leibniz University Hannover, Hannover, Germany}
\affil[2]{ TIB—Leibniz Information Centre for Science and Technology, Hannover, Germany}

\affil[*]{Correspondence: gollam.rabby@l3s.de}

\begin{document}

\maketitle

\section*{SUMMARY}

Mathematical reasoning creates significant challenges for large language models (LLMs). To improve the mathematical reasoning of LLMs, we propose Monte Carlo Self-Refine Tree (MC-NEST), an enhancement of Monte Carlo Tree, integrating LLM-based self-refinement and self-evaluation to improve decision-making in complex reasoning tasks.
MC-NEST balances exploration and exploitation using Upper Confidence Bound (UCT) scores and diverse selection policies. Iterative critique and refinement allow LLMs to reason more strategically.
Empirical results show that MC-NEST with an Importance Sampling Policy significantly increases GPT-4o performance, achieving the highest pass@1 on Olympic-level benchmarks. Specifically, MC-NEST attains a pass@1 of 38.6 on AIME and 12.6 on MathOdyssey.
The solution quality for MC-NEST with GPT-4o and Phi-3-mini reaches 84.0\% and 82.08\%, respectively, indicating robust consistency across LLMs.
MC-NEST demonstrates strong performance across Algebra, Geometry, and Number Theory, benefiting from its ability to manage abstraction, logical deduction, and multi-step reasoning—core to mathematical problem solving.

\section*{KEYWORDS}


Mathematical Reasoning, Large Language Models, Monte Carlo Self-Refining Tree

\section*{Introduction}
Large Language Models (LLMs) have demonstrated significant progress in mathematical problem-solving benchmarks such as GSM8K~\cite{cobbe2021training} and MATH~\cite{DBLP:conf/nips/HendrycksBKABTS21}. 
However, addressing complex mathematical reasoning tasks remains a critical challenge, particularly in Olympiad-level mathematics. 
These tasks require computational precision and advanced strategic reasoning—areas where current LLMs often fail. 
To address these limitations, recent efforts have explored enhancing structured reasoning through methods like Chain-of-Thought (CoT) prompting~\cite{fu2022complexity, wei2022chain} and Monte Carlo Tree Search (MCTS)~\cite{feng2023alphazero, pitanov2023monte}. 
While these approaches have progressed, they struggle with balancing exploration and exploitation, particularly in multi-step decision-making tasks~\cite{du2023improving, luo2023wizardmath}.

To solve this problem, we introduce the \textit{MC-NEST -- Monte Carlo Self-Refine Tree}, a method that extends the Monte Carlo Tree Self-Refine (MCTSr) method~\cite{zhang2024accessing} by incorporating probability distribution strategies and self-refinement mechanisms.
MC-NEST effectively balances search and refinement in the tree, enabling LLMs to tackle complex mathematical reasoning pathways with robustness. 
Key contributions of this work include: 
\begin{itemize}
    \item \textit{Probability Distribution Integration:} MC-NEST utilizes probability distribution to prevent biases on sub-optimal solutions by ensuring equitable exploration of solution paths.
    \item \textit{Dynamic Decision Policies:} Diverse strategies, such as Importance Sampling and Pairwise Importance Sampling, allow adaptive navigation across problem landscapes. 
    \item \textit{Iterative Self-Refinement:} Through iterative self-evaluation and refinement, MC-NEST improves LLM outputs for complex mathematical reasoning tasks.
\end{itemize}

In our experiments, we consider Zero-Shot Chain-of-Thought (ZSCoT)~\cite{kojima2022large} as a baseline, achieving a pass@1 score of 33.3. In comparison, MC-NEST demonstrates superior performance on the AIME dataset~\cite{parvez_zamil_gollam_rabby_2024}, attaining a pass@1 score of 38.6 and successfully solving 58 out of 150 complex mathematical problems. On the MathOdyssey dataset~\cite{DBLP:journals/corr/abs-2406-18321}, MC-NEST achieves a pass@1 score of 12.6, solving 19 out of 150 problems. These results underscore MC-NEST's robustness and effectiveness, particularly when the mathematical problem is complete and clearly described, where it outperforms the baseline in both dimensions of problem-solving.

\begin{figure*}[t]
    \centering
    \begin{minipage}[b]{0.45\textwidth}
        \includegraphics[width=\textwidth]{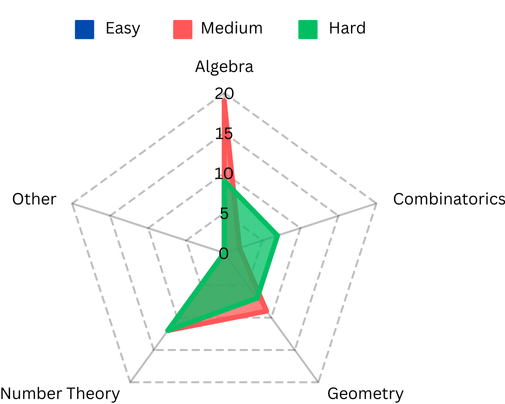}
        \caption{Solving different problems in different difficulty distributions across mathematical domains using MC-NEST.}
        \label{fig:figure1}
    \end{minipage}
    \hfill
    \begin{minipage}[b]{0.45\textwidth}
        \includegraphics[width=\textwidth]{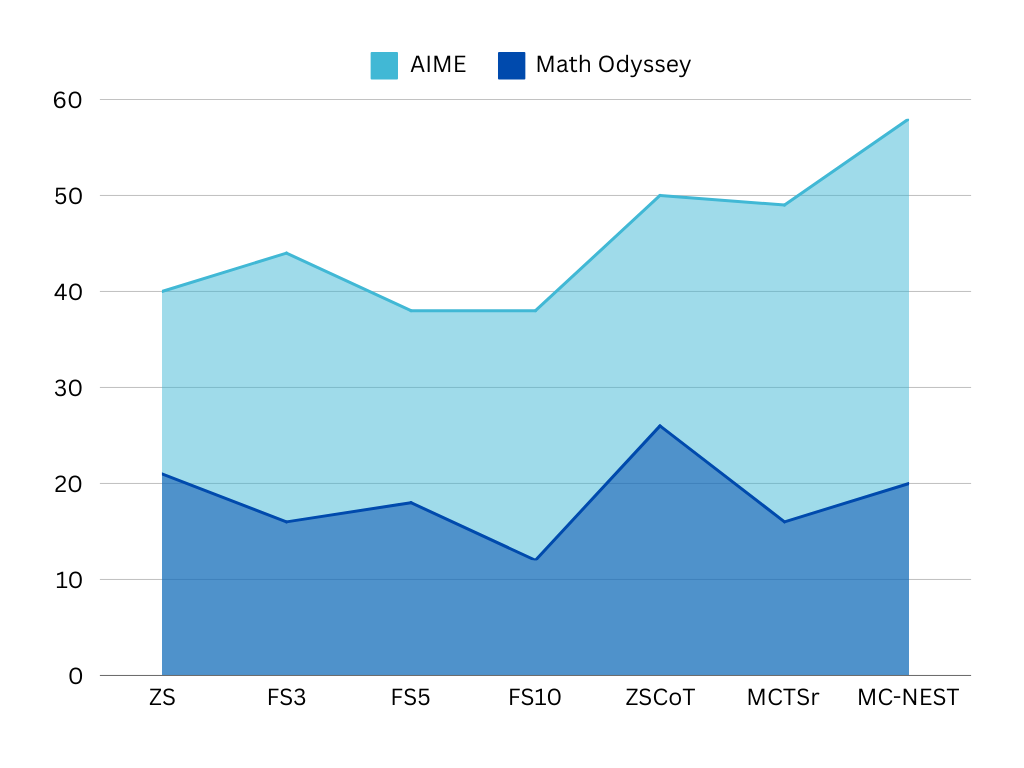}
        \caption{Comparison of problem-solving across different prompting techniques, MCTSr and MC-NEST with GPT-4o for AIME and MathOdyssey Datasets.}
        \label{fig:figure2}
    \end{minipage}
\end{figure*}

        \begin{figure*}[ht]
            \centering
            \includegraphics[height=7cm, width=\textwidth]{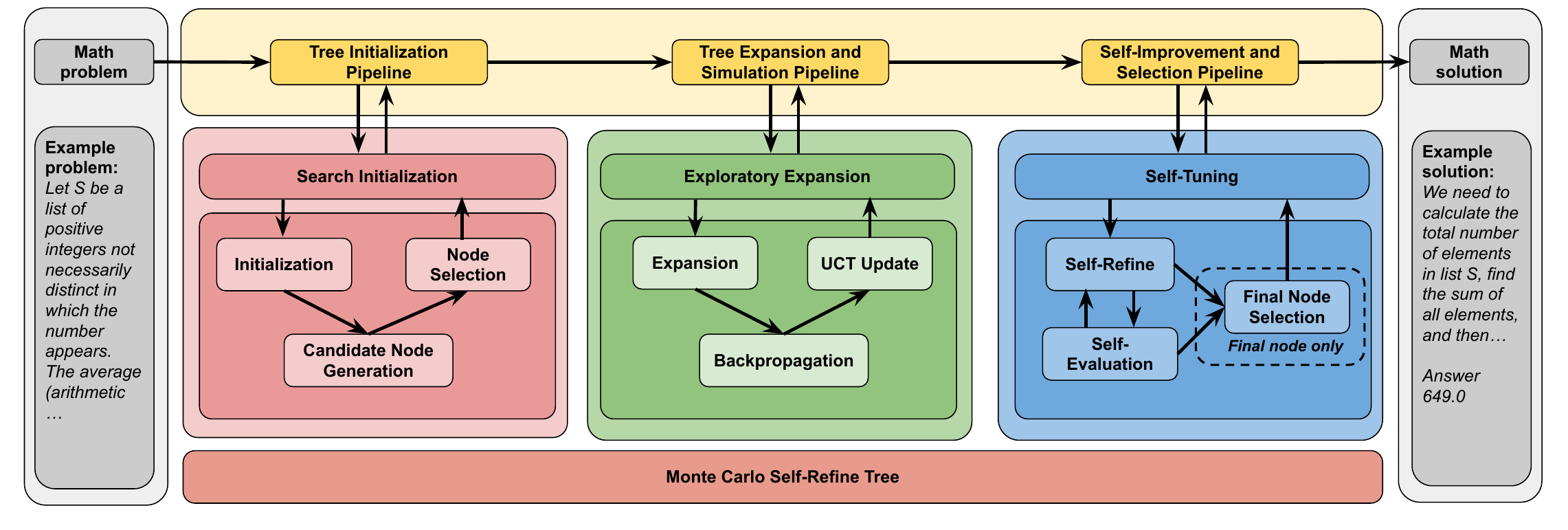}
            \caption{Architecture of the Monte Carlo Self-Refine Tree (MC-NEST) framework, showing the three-stage pipeline: (1) Tree Initialization (problem embedding and search space setup), (2) Tree Expansion and Simulation (Monte Carlo-based exploration with UCT backpropagation), and (3) Self-Improvement (solution refinement through self-tuning and evaluation).}
            \label{fig:methodology_overview}
        \end{figure*}

\section*{Related Work}

\paragraph{Improving Mathematical Reasoning Ability of LLMs}

Mathematical reasoning creates significant challenges for LLMs~\cite{team2023gemini, openai2023gpt, touvron2023llama}, serving as a key benchmark for evaluating their reasoning capabilities. Also, LLMs are outstanding at simple tasks but struggle with complex mathematical reasoning, raising concerns about their ability to generalize beyond basic problem-solving. Different prompting techniques, such as the chain-of-thought (CoT) proposed by Fu et al.~\cite{fu2022complexity} and Wei et al.~\cite{wei2022chain}, address these limitations by guiding LLMs for step-by-step reasoning. These prompting approaches enhance performance without tuning LLM parameters, improving pass@1 and boosting trust in LLM outputs. Advancing such approaches is critical for refining LLMs' reasoning capabilities and expanding their utility across complex domains.

\paragraph{Monte Carlo Search Tree}

Monte Carlo Tree Search (MCTS) has become a foundation in reinforcement learning (RL), powering breakthroughs like AlphaGo~\cite{silver2016mastering} and AlphaGo Zero~\cite{silver2017mastering}, where it was combined with deep RL techniques for exceptional performance in strategic decision-making~\cite{swiechowski2023monte}. In the context of LLMs, tree-based planning methods such as Tree-of-Thought~\cite{DBLP:conf/nips/YaoYZS00N23}, Reasoning-via-Planning~\cite{hao2023reasoning}, and inference-time MCTS~\cite{feng2023alphazero} enhance reasoning by exploring possible outputs systematically. Applications of MCTS extend beyond RL, including Multi-Agent Path-finding~\cite{pitanov2023monte}, Train Timetabling~\cite{yang2023integrated}, SAT problems~\cite{li2023general}, and robotics via PhyPlan~\cite{vagadia2024phyplan}. In LLMs, MCTS has shown promise in addressing challenges in multi-step reasoning. Approaches like AlphaMath~\cite{chen2024alphamath} and lightweight energy-based MCTS~\cite{xu2023no} improve mathematical reasoning, while the Monte Carlo Tree Self-Refine (MCTSr)~\cite{zhang2024accessing} integrates LLMs with systematic exploration and heuristic self-refinement. MCTSr demonstrates significant success across datasets, including Olympiad-level benchmarks, advancing LLM-driven applications in strategic reasoning.

\begin{algorithm*}
\caption{MC-NEST for Complex Mathematical Reasoning}
\label{mc-nest}
\begin{algorithmic}[1]
    \STATE \textbf{Input:} Problem $P$, initial answer $A_0$, number of iterations $N$, LLM
    \STATE \textbf{Output:} Refined Answer $A$
    
    \STATE Initialize the root node $n_0$ with initial answer $A_0$
    \FOR{$i = 1$ to $N$}
        \STATE \textbf{Selection:} Select a node $n_i$ using UCT and probability distribution strategy
        \STATE \textbf{Expansion:} If $n_i$ is not fully expanded, expand by adding a new child node $n_c$
        \STATE \textbf{Self-Refine:}
        \STATE Generate critique $C_i = \text{LLM}(\texttt{CritiquePrompt}(P, A_i))$
        \STATE Refine the answer $A_{i+1} = \text{LLM}(\texttt{RefinePrompt}(P, A_i, C_i))$
        \STATE Add $n_c$ with refined answer $A_{i+1}$ to the children of $n_i$
        \STATE \textbf{Self-Evaluate:}
        \STATE Evaluate the refined answer $R_i = \text{LLM}(\texttt{EvaluatePrompt}(P, A_{i+1}))$
        \IF{$R_i$ exceeds reward limit}
            \STATE Apply penalty $R_i = R_i - \texttt{penalty}$
        \ENDIF
        \STATE \textbf{Backpropagation:}
        \STATE Update $Q$ values and visit counts for the parent nodes of $n_c$
        \STATE Backpropagate the reward $R_i$ up to the root node
    \ENDFOR
\STATE \textbf{Return} the answer $A$ from the node with the highest $Q$ value
\end{algorithmic}
\end{algorithm*}

\paragraph{LLM As An Evaluator}

Recent advancements in evaluating LLMs have introduced innovative methods to enhance LLM assessment. One notable approach leverages pre-trained LLMs to simulate human evaluators, with systems like AlpacaFarm~\cite{dubois2024alpacafarm} utilizing preference scores between LLM outputs. Different studies investigate GPT-4’s capacity as like as human preferences~\cite{goli2024frontiers, mitchell2023comparing}. Additionally, LLMBar~\cite{DBLP:conf/iclr/ZengYG0G024} provides a systematic evaluation framework for instruction-following tasks. These studies consistently highlight GPT-4's reliability, forming the foundation for the quality of MC-NEST-solved complex Olympiad-level mathematical problem solutions in comparison to human-generated solutions. 



\section*{METHOD: Monte Carlo Self-Refine Tree}

We introduce MC-NEST, a novel Monte Carlo Tree Search-based method that integrates probability distribution strategies and LLM-driven self-refinement to enhance complex mathematical reasoning. 
MC-NEST employs an iterative process, selecting optimal nodes using Upper Confidence Bounds (UCT) augmented with probability distribution and refining solutions through LLM-generated critiques. 
Node evaluations leverage an LLM-based self-evaluation mechanism, which scores candidate solutions and backpropagates rewards to guide convergence toward optimal search paths. 
The method encompasses different stages—Initialization, Candidate Node Generation, Node Selection, Expansion, Backpropagation, UCT Update, Self-Evaluation, Self-Refine, and Final Node Selection, detailed in Algorithm~\ref{mc-nest} and visualized in \autoref{fig:methodology_overview}. This methodology systematically improves pass@1 through structured exploration and refinement.

\subsection{Initialization}
\label{subsub:Initialization}

The initialization phase of the MC-NEST establishes the root node, which determines the initial direction of the search. This can be achieved using either a ZSCoT strategy or a predefined dummy answer. In the ZSCoT approach, the root node is initialized with the output of a pre-trained LLM, producing an initial solution to the problem based on the problem instance \( p \) without prior search history:
    
     \[\text{root} = \text{Node}(\text{answer} = \text{ZeroShotCoT\_LLM}(p)),\]

where \( \text{ZeroShotCoT\_LLM}(p) \) represents the LLM's ZSCoT reasoning output. This initialization leverages the LLM's pre-trained knowledge to provide a strong starting point for further refinements by MC-NEST. Alternatively, when no LLM-based prior knowledge is available, the root node can be initialized with a neutral dummy answer (e.g., "I don't know."), allowing the method to start from a minimal baseline.

\subsection{Candidate Node Generation}
\label{subsub:Candidate_Node_Generation}

After initializing the root node, the method constructs a search tree with the root node serving as a starting point.
Child nodes are added to a parent node using a prompt (cf. \autoref{refinement-prompt}), applying self-refine and self-evaluation on the parent node to improve the solution.

In contrast to standard MCTS, which typically applies the UCT criterion recursively during descent, the MC-NEST method performs a breadth-first search (BFS) traversal in the node selection phase.
Starting from the root node, BFS expands to child nodes to generate a candidate set for further expansion.
Each node is evaluated as a candidate node based on specific criteria: it must not have reached the maximum allowed number of children, and none of its children should possess a higher quality score \( Q \).
This BFS-based candidate generation serves as a preparatory filtering step, ensuring a wider and more balanced exploration of the search space at early stages, which is particularly important given the high variance in solution quality typical of LLM-generated outputs.

If no eligible candidates are found, the method defaults to focusing on the root node, facilitating iterative improvement of the overall solution.
By combining BFS for breadth-aware candidate filtering with subsequent UCT-based selection, MC-NEST effectively balances exploration and exploitation, guiding the search towards optimal solutions.

\subsection{Probability Distribution Strategy for Node Selection}
\label{subsub:Nash_Equilibrium_Strategy_for_Node_Selection}

Following the BFS-based candidate generation phase, the node selection process identifies the most promising candidate node for further evaluation by combining UCT scores (\autoref{sub:UCT_update}) with various selection policies, refined by probability distribution strategies.

A node is considered fully expanded if it has reached the maximum number of children or if any of its children possess a reward \( Q \) greater than or equal to the current node's.
To formalize this process, consider a set of candidate nodes \( A = \{a_1, a_2, \dots, a_n\} \) generated via BFS.

The probability distribution strategy initially assigns a uniform probability to each candidate node to ensure equal treatment in the absence of additional information.
When no strong preference is established through UCT scores, all non-expanded nodes continue with an equal probability of exploration.
The probability of selecting a node \( a_i \) is defined as:

\[
\pi(a_i) = \frac{1}{n}, \quad \forall i = 1, 2, \dots, n
\]

where \( n \) is the total number of candidate nodes, and the probabilities satisfy \( \sum_{i=1}^{n} \pi(a_i) = 1 \).

This uniform distribution is then integrated with the UCT scores to guide final node selection.
Three selection policies are employed:

\paragraph{Greedy Policy} selects the node with the highest combined UCT score and probability distribution: 
\[
\text{Score}(i) = UCT(i) + \pi(a_i)
\]
The node with the highest score is selected:
\[
i^* = \arg\max_i \left[ \text{Score}(i) \right]
\]

\paragraph{Importance Sampling Policy} assigns weights based on UCT scores and probability distribution and selects nodes probabilistically:
\[
\text{Weight}(i) = UCT(i) \times \pi(a_i)
\]
The node is then sampled according to these weights:
\[
i^* = \text{random\_choice}(C, \text{weights} = \{\text{Weight}(i)\})
\]
While this involves weighted sampling, it differs from traditional importance sampling, which estimates expectations under a target distribution using a proposal distribution.

\paragraph{Pairwise Importance Sampling Policy} evaluates pairs of nodes based on the absolute difference in their UCT scores and their probability distribution weights:
\[
\text{Pair\_Weight}(i,j) = |\text{UCT}(i) - \text{UCT}(j)| \times \pi(a_i) \times \pi(a_j)
\]
The node with the higher UCT score in the pair is selected:
\[
i^* = \arg\max \left( \text{UCT}(i), \text{UCT}(j) \right)
\]

This two-stage selection process—initial BFS candidate filtering followed by UCT-guided selection integrated with probabilistic policies—ensures both breadth-aware exploration and exploitation of high-quality nodes, enhancing the stability and effectiveness of the MC-NEST search algorithm.

\subsection{Upper Confidence Bound (UCT) Update}
\label{sub:UCT_update}

The UCT update in MC-NEST optimally guides the node selection by computing the UCT value for node \(i\) as:
    
   \[UCT(i) = Q(i) + C \sqrt{\frac{\ln(N_{\text{parent}})}{N(i) + \epsilon}}\]
    
where \( Q(i) \) is the current reward value of node \(i\), \( C \) is the exploration constant controlling the trade-off, \( N_{\text{parent}} \) is the number of visits to the parent node of \(i\), \( N(i) \) is the number of visits to node \(i\), and \( \epsilon \) is a small constant to avoid division by zero. This UCT score suggests nodes with higher rewards for exploitation while encouraging exploration of less-visited nodes. To synchronize the UCT update with the probability distribution strategy introduced earlier, the node selection score is adjusted by adding a uniform probability, resulting in the updated score:
    
    \[\text{UCT}(i) = Q(i) + C \sqrt{\frac{\ln(N_{\text{parent}})}{N(i) + \epsilon}} + \frac{1}{n}\]

where \( \pi(a_i) = \frac{1}{n} \) represents the uniform probability over \(n\) candidate nodes. This adjusted score ensures a balance between exploration, exploitation, and fairness across nodes. The node with the highest score, \(i^* = \arg\max_i [\text{Score}(i)]\), is selected for further exploration or refinement, or as a final node. This integration of the UCT update with the probability distribution strategy ensures a robust search, balancing known rewards with the need for continued exploration across all potential solutions.

\subsection{Expansion}
\label{sub:Expansion}

The expansion step in MC-NEST follows the node selection and involves generating a new child node to further explore the search tree. After selecting a node \( n_s \), the method refines the current solution stored in \( n_s \) to create a new child node \( n_c \), thereby advancing the exploration and refinement process. Formally, this can be represented as: \smash{$n_c = \text{SelfRefine}(n_s)$}, where \( \text{SelfRefine}(n_s) \) denotes the refinement process that improves the current answer at node \( n_s \). This refinement is achieved by critiquing the current answer and generating a proposed improvement, which is then stored in the new child node \( n_c \): \smash{$n_s \text{.children} \gets n_s \text{.children} \cup \{n_c\}$}. The self-refinement process utilizes the same LLM to critique the current answer \( a_s \) at node \( n_s \). The critique is generated as: \smash{$\text{Critique}(a_s) = \text{LLMCritique}(p, a_s)$}, where \( p \) is the problem instance and \( a_s \) is the answer at node \( n_s \). Based on this critique, the answer is refined to produce an improved solution \( a_c \): \smash{$a_c = \text{RefineAnswer}(p, a_s, \text{Critique}(a_s))$}. The refined answer \( a_c \) is then stored as child node \( n_c \), which is linked as a child of \( n_s \). This expansion process allows MC-NEST to iteratively improve and refine solutions, ensuring that the search progresses toward better solutions in a structured and systematic manner.

\subsection{Backpropagation}
\label{sub:Backpropagation}

The backpropagation step in MC-NEST updates the quality score (\( Q \)) and visit count of each node from the newly expanded child node back up to the root. This step ensures that information from deeper explorations in the tree is propagated upwards, influencing future decisions at higher levels. After the expansion phase, let the newly created child node be denoted as \( n_c \) and its parent node as \( n_p \). The backpropagation process recursively updates the quality score and visit count of each parent node, starting from \( n_p \) and moving upward to the root. For each parent node \( n_p \), the quality score \( Q(n_p) \) is updated based on the maximum \( Q \) value among its children. This update ensures that the best information from deeper nodes influences higher-level nodes, guiding the search process. The updated quality score for a parent node \( n_p \) is given by:
\[Q(n_p) = (Q(n_p) + \max(Q(n_c)))/{2}\]
where \( Q(n_p) \) is the current quality score of the parent node and \( Q(n_c) \) is the quality score of the child node \( n_c \). This averaging mechanism ensures the parent node’s quality score reflects both its current performance and the best-performing child node, balancing the exploitation of known information with exploration of new areas. In addition to updating the quality score, the visit count for each parent node is incremented by one during backpropagation: \smash{$\text{Visit}(n_p) = \text{Visit}(n_p) + 1$}. This ensures that each parent node’s visit count accurately reflects how often it has been involved in the backpropagation process. A higher visit count indicates greater confidence in the information coming from that node. The backpropagation process continues recursively from the newly expanded child node \( n_c \), through its parent \( n_p \), and so on, until it reaches the root node. This recursive update ensures that information from deep within the search tree influences the root node’s quality score and visit count, impacting future node selections during the MC-NEST search process.

\begin{table*}
    \centering
    \small
    \begin{tabular}{l|c|c|c|c|c|c|c}
    \toprule
    \multirow{2}{*}{\textbf{Domain}} & \multicolumn{3}{c|}{\textbf{AIME}} & \multicolumn{4}{c}{\textbf{MathOdyssey}} \\ 
    \cline{2-8}
    & \textbf{Hard} & \textbf{Medium} & \textbf{Total} & \textbf{Easy} & \textbf{Hard} & \textbf{Medium} & \textbf{Total} \\
    \hline
    Algebra & 10 & 35 & 45 & 3 & 15 & 19 & 37 \\
    Combinatorics & 3 & 17 & 20 & 0 & 15 & 17 & 32 \\
    Geometry & 16 & 27 & 43 & 0 & 17 & 11 & 28 \\
    Number Theory & 10 & 28 & 38 & 0 & 25 & 24 & 49 \\
    Other & 1 & 3 & 4 & 0 & 2 & 2 & 4 \\ \hline
    Total & 40 & 110 & 150 & 3 & 74 & 73 & 150 \\
    \hline
    \end{tabular}
    \caption{Comparison of difficulty levels across domains for AIME and MathOdyssey with total counts per domain.}
    \label{tab:combined_difficulty}
\end{table*}

\subsection{Self-Refine}
\label{sub:Self_Refine}

The self-refine approach in the MC-NEST method iteratively improves a candidate solution using a critique-and-refinement process based on feedback from the LLM~\cite{DBLP:journals/corr/abs-2310-04815, DBLP:journals/corr/abs-2408-12315, DBLP:journals/corr/abs-2206-05802}. This method is designed to refine an answer at each node in the tree by generating critiques and then enhancing the answer based on these critiques. Let \( n \) represent the current node in the search tree, which contains a candidate answer \( A_n \). To refine this answer, a critique is first generated using a prompt directed to the LLM. The LLM produces a critique of the current answer, identifying potential weaknesses or areas of improvement. The prompt used for generating the critique consists of the problem \( P \) and the current answer \( A_n \), formatted as follows: \smash{$\text{CritiquePrompt} = \left( \text{problem} + P + \text{current\_answer} + A_n \right)$}.
    The LLM takes this prompt and generates the critique \( C_n \):  \[\smash{C_n = \text{LLM}\left( \text{CritiquePrompt} \right)}\]
Once the critique \( C_n \) is generated, it is utilized to refine the current answer \( A_n \) into a new, improved answer \( A_{n+1} \). This is achieved through another prompt to the LLM, which incorporates the problem \( P \), the current answer \( A_n \), and the generated critique \( C_n \):
        \[\text{RefinePrompt} = \left( P  + A_n +  C_n \right)\] 
    The LLM refines the answer, producing a more accurate or improved answer \[ A_{n+1} : A_{n+1} = \text{LLM}\left( \text{RefinePrompt} \right)\]
    The newly refined answer is returned and stored in a new node \( n+1 \), which is added as a child of the current node \( n \) in the search tree. This process constructs a new node with a refined answer, continuously enhancing the search for optimal solutions within the MC-NEST method.

    \subsection{Self-Evaluation}
    \label{sub:Self_Evaluation}

    The self-evaluate approach in MC-NEST assesses a candidate answer at a given node by assigning a reward based on how well the answer solves the problem. This reward is then utilized to update the node's statistics, including the total reward and the visit count, ensuring that the node’s quality is improved through evaluation. Let \( n \) be a node in the search tree, where \( n \) contains a candidate answer \( A_n \). The self-evaluation process begins with determining the quality of this answer using a reward function implemented in the evaluate answer method. The reward function assesses the answer's quality by querying an LLM with the problem \( P \) and the answer \( A_n \), as follows:    
    \[R_n = \text{LLM}\left( \texttt{EvaluatePrompt}(P, A_n) \right)\]
    Here, \( R_n \) represents the reward assigned to the answer \( A_n \), which is an integer reflecting how well the answer addresses the problem \( P \). This value quantifies the node's quality. The reward assigned to a node is subject to constraints defined by MC-NEST. Specifically, if the reward exceeds a predefined threshold known as the reward limit, a penalty is applied to prevent excessively high rewards from disproportionately influencing the search process. The penalized reward \( \tilde{R}_n \) is computed as follows:
    
    \[
    \tilde{R}_n = 
    \begin{cases} 
    R_n & \text{if } R_n \leq \text{$R_n$\_limit} \\
    R_n - \text{excess\_reward\_penalty} & \text{if } R_n > \text{$R_n$\_limit}
    \end{cases}
    \]
    
    where the reward limit is the upper bound on the reward, and the excess reward penalty is the penalty applied when the reward exceeds this limit. Once the reward \( \tilde{R}_n \) has been determined, the node's statistics are updated. These updates include adding the reward to the node's total reward and incrementing the node's visit count. Formally, for each node \( n \), the following operations are performed: \smash{$\text{TotalReward}_n = \text{TotalReward}_n + \tilde{R}_n$} and \smash{$\text{VisitCount}_n = \text{VisitCount}_n + 1$}. Consequently, the node’s reward and visit counts are updated to reflect the evaluation results, contributing to improved decision-making.

\section*{Experiments}

In our experiments we utilized ZSCoT prompting as our base prompting style with GPT-4o~\cite{achiam2023gpt}, Mistral (7B)~\cite{DBLP:journals/corr/abs-2310-06825} and Phi-3-mini (3.8B)~\cite{abdin2024phi} LLM.

    \subsection{Olympiad-level Dataset}
    \paragraph{Data Categories}
        In this experiment, we utilized GPT-4o to systematically classify the math problems from the AIME and MathOdyssey datasets into distinct mathematical domains and difficulty levels. The primary aim was to categorize each problem into one of the following mathematical fields: \textit{Number Theory}, \textit{Geometry}, \textit{Algebra}, \textit{Combinatorics}, or \textit{Other}. Additionally, each problem was assigned a difficulty level of either \textit{Easy}, \textit{Medium}, or \textit{Hard}. To validate the performance of the LLM, the classifications were cross-checked through human-based verification. Each mathematical problem's assigned field and difficulty level provided an additional layer of investigation. This human-based verification process increases the reliability of the LLM's classifications. A detailed discussion of the results, including the LLM's performance in both mathematical field and difficulty classification, is presented in \autoref{sec:Results}.
        
    \paragraph{Dataset} 
        
        In total, we utilized 300 complex Olympiad-level mathematical problems from the AIME and MathOdyssey datasets covering different domains such as Algebra, Combinatorics, Geometry, and Number Theory. The AIME dataset comprises 150 problems, including 40 hard and 110 medium problems, while MathOdyssey also includes 150 problems, with 74 hard, 73 medium, and 3 easy problems. For each problem, at least one human-generated solution was provided to ensure correctness and quality. Since this dataset represents a subset of complex Olympiad-level problems, we plan to expand it in future work. Detailed statistics are provided in~\autoref{tab:combined_difficulty}.

    \subsection{Prompt Engineering}

    Zero-shot (ZS) prompting~\cite{wei2021finetuned} enables LLMs to perform tasks using direct instructions without examples or prior context. Effective for simple tasks like text classification, its limitations appear in complex scenarios requiring detailed understanding or multi-step reasoning.
    Few-shot (FS) prompting~\cite{brown2020language} improves upon this by incorporating a few task-specific examples within the prompt, enhancing in-context learning and achieving better performance.
    However, FS faces challenges in handling advanced reasoning tasks, such as logic or arithmetic problems.
    Retrieval-Augmented Generation (RAG)~\cite{lewis2020retrieval} addresses these issues by dynamically retrieving semantically relevant examples beyond the set of utilized datasets using tools like Facebook AI Similarity Search (FAISS)~\cite{douze2024faiss,johnson2019billion}.
    With embeddings from "SFR-Embedding Mistral"~\cite{meng2024sfrembedding}.
    RAG integrates these examples into the prompt, improving contextual alignment and task accuracy. 
    ZSCoT prompting further advances reasoning capabilities by guiding LLMs through step-by-step thinking, effectively solving multi-step problems.
    ZSCoT shows strong performance on complex tasks with large LLMs.
    However, prompt construction in ZSCoT is resource-intensive, and ambiguity in intermediate steps can lead to errors.
    Within all of these prompting approaches, ZSCoT offers significant potential for improving LLM reasoning with careful application and iterative refinement.

    \subsection{Evaluation Matrix}
            
        \paragraph{Pass@1}    
        Pass@1 served as a critical evaluation metric to measure the performance of LLMs in solving complex Olympiad-level mathematics problems~\cite{openai2024learning}.
        Pass@1 is defined as the ratio of problems correctly solved on the first attempt to the total number of problems attempted by the LLM. 
        Formally, if the total number of problems is denoted by $N$, and the number of correctly solved problems on the first attempt by $T_{\text{correct}}$, then Pass@1 is calculated as:  
        \[
        \smash{
        \text{Pass@1} = \frac{T_{\text{correct}}}{N}
        }
        \]
        A higher Pass@1 score indicates a superior performance of the LLM, reflecting its ability to generate accurate solutions on the first try.
        This metric evaluates the logical precision and mathematical reasoning of LLMs when tackling Olympiad-level mathematics, which often demands high rigor and complexity.

        \paragraph{Rollout Selection} The rollout performance was evaluated based on the number of correctly solved problems out of 150 complex mathematical problems from each of the AIME and MathOdyssey datasets.
        For each problem, rollouts were performed at intervals of 4, 8, 12, 16, 20, 24, 28, 32, and 36 iterations.
        This approach was used to select the optimal rollout strategy for MC-NEST across different LLMs.
        The output for each complex math problem was crucial for selecting the best rollout, and all results were thoroughly verified. Additionally, human evaluators cross-checked the responses generated by the LLMs to ensure their correctness.

    \subsection{LLM Selection}

    For the MC-NEST evaluation of Olympiad-level math problems, we select GPT-4o, Mistral (7B), and Phi-3 (3.8B) based on their strengths and alignment with the experimental goals.
    GPT-4o serves as a robust general-purpose LLM baseline due to its proven reasoning capabilities~\cite{DBLP:journals/corr/abs-2303-08774, openai2024learning}.
    Mistral (7B) is chosen for its efficiency and high performance for other reasoning tasks relative to its parameter size, providing insights into scaling trade-offs~\cite{DBLP:journals/corr/abs-2310-06825}.
    Phi-3-mini (3.8B) is selected for its optimization toward mathematical reasoning, making it well-suited for solving domain-specific problems~\cite{DBLP:journals/corr/abs-2404-14219}.
    These selections allow for a diverse evaluation across different architectures, sizes, and training methodologies, ensuring meaningful insights into the efficacy of MC-NEST in solving complex mathematical problems.

    \subsection{Quality Checker}

    \paragraph{Key Points as Ground-Truths}

        We utilized human-verified ground-truth solutions for each mathematical problem to identify the key attributes necessary for a comprehensive quality assessment.
        These attributes include (1) \textit{Completeness}: assessing whether all necessary steps are included and logically connected; (2) \textit{Clarity}: ensuring that the explanation and notation are easy to follow and unambiguous; (3) \textit{Optimality}: evaluating the efficiency of the solution; and (4) \textit{Mathematical Rigor}: verifying that all hypothesis, justifications, and definitions are explicitly and appropriately stated, which collectively define the criteria for evaluating the quality of a solution.
        By focusing on these attributes, it became feasible to assess not only the correctness of a solution but also the quality of solutions generated using the MC-NEST approach.

    \paragraph{Human-LLM Based Quality Check}
    For the quality check, the December 2024 version of GPT-4o is prompted to evaluate the quality of each complex problem solution generated by the MC-NEST approach against one of the human-proven solutions.
    We also generate responses of up to 1,000 tokens to ensure comprehensive evaluations.
    We run this quality check approach multiple times to ensure the consistency of the evaluations and calculate the average to provide a reliable assessment.

\begin{table}
    \centering
    \small
        \begin{tabular}{c|c|c|c|c|c|c|c|c|c|c|c|c}
            \toprule
            \textbf{LLM} & \textbf{Size} & \textbf{Method} & \textbf{Node Selection} & \textbf{4} & \textbf{8} & \textbf{12} & \textbf{16} & \textbf{20} & \textbf{24} & \textbf{28} & \textbf{32} & \textbf{36} \\
            \midrule
            & & MCTSr & Gre  & 46 & 46 & 38 & 42 & 33 & 34 & 46 & 36 & 37 \\
            & & MCTSr & IS  & \textbf{49} & 42 & 44 & 36 & 40 & 38 & 35 & 32 & 40 \\
            & & MCTSr & PIS  & 42 & 39 & 36 & 38 & 32 & 38 & 39 & 34 & 32 \\
            GPT4o & - & MC-NEST & Gre  & 42 & 45 & 38 & 38 & 38 & 39 & 35 & 32 & 39 \\
            & & MC-NEST & IS  & \textbf{58} & 37 & 33 & 37 & 30 & 34 & 30 & 34 & 31 \\
            & & MC-NEST & PIS  & 50 & 35 & 41 & 37 & 44 & 28 & 35 & 30 & 28 \\
            \midrule
            & & MCTSr & Gre  & \textbf{11} & 1 & - & - & - & - & - & - & - \\
            & & MCTSr & IS  & 1 & - & - & - & - & - & - & - & -\\
            & & MCTSr & PIS  & 8 & 1 & - & - &  - & - & - & - & -\\
            Phi-3-mini & & MC-NEST & Gre  & 6 & 3 & - & - & - & - & - & - & -\\
            & & MC-NEST & IS  & \textbf{12} & 2 & - & - & - & - & - & - & -\\
            & & MC-NEST & PIS  & 9 &  7 & 1 & 2 & - & - & - & - & -\\
            \midrule
            & & MCTSr & Gre  & - & - & - & - & - & - & - & - & - \\
            & & MCTSr & IS  & - & - & - & - & - & - & - & - & - \\
            & & MCTSr & PIS  & - & - & - & - & - & - & - & - & - \\
           Mistral & 7B & MC-NEST & Gre  & - & - & - & - & - & - & - & - & - \\
            & & MC-NEST & IS  & - & - & - & - & - & - & - & - & - \\
            & & MC-NEST & PIS  & - & - & - & - & - & - & - & - & - \\
            \bottomrule
        \end{tabular}
        \caption{Different rollouts for different node selection methods (for selecting the best rollout) with MC-NEST in 150 Olympiad level (AIME dataset) math problems using different LLMs; Gre = Greedy; IS = Importance Sampling; PIS = Pairwise Importance Sampling.}
        \label{tab:AIME_rollout_result}
\end{table}

\begin{table}
    \centering
    \small
        \begin{tabular}{c|c|c|c|c|c|c|c|c|c|c|c|c}
            \toprule
            \textbf{LLM} & \textbf{Size} & \textbf{Method} & \textbf{Node Selection} & \textbf{4} & \textbf{8} & \textbf{12} & \textbf{16} & \textbf{20} & \textbf{24} & \textbf{28} & \textbf{32} & \textbf{36} \\
            \midrule
            & & MCTSr & Gre & \textbf{16} & 12 & 11 & 9 & 13 & 11 & 12 & 8 & 14 \\
            & & MCTSr  & IS & \textbf{16} & 13 & 8 & 8 & 9 & 12 & 7 & 9 & 10 \\
            & & MCTSr & PIS & 13 & 15 & 10 & 14 & 8 & 7 & 9 & 7 & 13 \\
            GPT4o & - & MC-NEST & Gre & 11 & 9 & 10 & 11 & 10 & 10 & 8 & 10 & 5  \\
            & & MC-NEST & IS & 18 & 12 & 12 & 14 & 13 & \textbf{20} & 10 & 17 & 8 \\
            & & MC-NEST & PIS & 16 & 14 & 10 & 9  & 11 & 7 & 12 & 11 & 9 \\
            \midrule
            & & MCTSr  & Gre & \textbf{3} & - & - & - & - & - & - & - & - \\
            & & MCTSr  & IS & 2 & - & - & - & - & - & - & - & -  \\
            & & MCTSr & PIS & 3 & - & - & - & - & - & - & - & -  \\
            Phi-3-mini & & MC-NEST  & Gre & \textbf{4} & - & - & - & - & - & - & - & - \\
            & & MC-NEST  & IS & 2 & - & - & - & - & - & - & - & -  \\
            & & MC-NEST  & PIS & 1 & - & - & - & - & - & - & - & -  \\
            \midrule
            & & MCTSr & Gre & - & - & - & - & - & - & - & - & - \\
            & & MCTSr & IS & - & - & - & - & - & - & - & - & - \\
            & & MCTSr & PIS & - & - & - & - & - & - & - & - & - \\
            Mistral & 7B & MC-NEST & Gre & - & - & - & - & - & - & - & - & - \\
            & & MC-NEST & IS & - & - & - & - & - & - & - & - & - \\
            & & MC-NEST & PIS & - & - & - & - & - & - & - & - & - \\
            \bottomrule
        \end{tabular}
        \caption{Different rollout for different node selection methods (for selecting the best rollout) with MC-NEST in 150 Olympiad level (MathOdyssey dataset) math problem using different LLMs.}
        \label{tab:math_oddesy_rollout_result}
\end{table}

\begin{table*}[ht]
    \centering
    \begin{tabular}{l|l|r|r|r|r|r|r|r|r}
    \hline
    \multirow{2}{*}{\textbf{Dataset}} & \multirow{2}{*}{\textbf{LLM}} & \multirow{2}{*}{\textbf{Size}} & \multicolumn{5}{c|}{\textbf{Prompts}} & \textbf{MCTSr} & \textbf{MC-NEST} \\ \cline{4-10}
     & & & \textbf{ZS} & \textbf{FS3} & \textbf{FS5} & \textbf{FS10} & \textbf{ZSCoT} & \textbf{} & \textbf{} \\ \hline
         & GPT-4o & - & 26.6 & 29.3 & 25.3 & 25.3 & 33.3  & 32.6 & \textbf{38.6} \\ 
    \textit{AIME} & Mistral & 7B & 1.3 & \textbf{2.6}  & \textbf{2.6} & \textbf{2.6} & 1.3 & - & - \\ 
         & Phi-3-mini & 3.8B & 4.6 & \textbf{8.0}  & 5.3 & 6.6 & 1.3 & 7.33 & 7.33 \\ \hline
         
         & GPT-4o & - & 14.0 & 10.6 & 12.0 & 8.0 & \textbf{16.6} & 10.6 & 13.3 \\ 
    \textit{MathOdyssey} & Mistral & 7B & 18.0 & \textbf{18.6}  & \textbf{18.6} & 0.6 & 13.3 & - & - \\ 
         & Phi-3-mini & 3.8B & 8.0 &  \textbf{14.0} & \textbf{14.0} & 4.6 & 8.6 & 2.0 & 2.0 \\ \hline
    
         \hline

    \end{tabular}
    \caption{Pass@1 comparison of different LLMs with various approaches on specific datasets. (ZS = Zero-Shot Prompting; FS = Few-Shot Prompting; ZSCoT = Zero-Shot Chain-of-Thought Prompting; MCTSr = Monte Carlo Tree Self-Refine (Rollout 4 with Importance Sampling Policy for AIME, Rollout 4 with Greesy Policy for MathOdyssey); MC-NEST = Monte Carlo Self-Refine Tree (Rollout 4 with Importance Sampling Policy for AIME, Rollout 24 with Importance Sampling Policy for MathOdyssey).}
    \label{tab:gpt4o_phi_3_com_performance_comparison}
\end{table*}

\begin{table*}
    \centering
    \small
    \begin{tabular}{p{1.2cm}|p{2.3cm}|p{1.5cm}|p{0.8cm}|p{2.2cm}|p{2.3cm}|p{2cm}|p{0.8cm}}
    \hline
    \textbf{Dataset} & \textbf{Domain}        & \textbf{Difficulty} & \textbf{Solve} & \textbf{Dataset} & \textbf{Domain}        & \textbf{Difficulty} & \textbf{Solve} \\ \hline
                 & Algebra                & Hard                & 6                    &          & Algebra                & Hard                & 3                    \\ \hline
                 & Algebra                & Medium              & 15                   &          & Algebra                & Medium              & 4                    \\ \hline
                 & Combinatorics          & Hard                & 5                    &          & Combinatorics          & Hard                & 2                    \\ \hline
    AIME             & Combinatorics          & Medium              & 2                    & MathOdyssey         & Combinatorics          & Medium              & 1                    \\ \hline
                 & Geometry               & Hard                & 6                    &          & Geometry               & Hard                & 1                    \\ \hline
                 & Geometry               & Medium              & 7                    &          & Geometry               & Medium              & 2                    \\ \hline
                 & Number Theory          & Hard                & 9                    &          & Number Theory          & Hard                & 3                    \\ \hline
                 & Number Theory          & Medium              & 8                    &          & Number Theory          & Medium              & 4                    \\ \hline
    \end{tabular}
    \caption{Domain Difficulty and solved with Datasets: AIME (MC-NEST - rollout 4) and mathodesy (MC-NEST - rollout 24)}
    \label{tab:domain_difficulty_solved}
\end{table*}

\section*{Results}
\label{sec:Results}

\textbf{Rollout Selection} Our rollout evaluation of MC-NEST and MCTSr across the AIME and MathOdyssey datasets reveals distinct performance patterns based on rollouts and node selection policies. On the AIME dataset, GPT-4o achieves its best performance with a rollout of 4 using the Importance Sampling policy, solving 58 problems, while on MathOdyssey, a rollout of 24 with the same policy yields the highest performance, solving 20 problems. For Phi-3-mini (3.8B), the optimal results on AIME are obtained with a rollout of 4 using Importance Sampling, solving 12 problems, whereas on MathOdyssey, a rollout of 4 with the Greedy policy solves 4 problems. Notably, Mistral (7B) fails to solve any problems across both datasets, highlighting its limitations in handling self-refine and self-evaluation for Olympiad-level mathematical tasks. These findings emphasize the critical role of customized rollouts and node selection strategies in maximizing the effectiveness of LLMs for complex problem-solving, as detailed in~\autoref{tab:AIME_rollout_result} and~\autoref{tab:math_oddesy_rollout_result}.

\textbf{Pass@1} Our detailed comparative analysis of MC-NEST against the GPT-4o, Mistral (7B), and Phi-3-mini (3.8B) across the two complex Olympiad-level mathematical datasets is presented in \autoref{tab:gpt4o_phi_3_com_performance_comparison}. 
The AIME dataset, known for its high-level problem-solving challenges, serves as a rigorous test for any LLM in this direction.
MC-NEST achieves a top pass@1 of 38.6, outperforming GPT-4o by a significant margin, whose pass@1 ranges from 25.3 to 33.3 depending on the prompting style.
Notably, the smaller LLMs Mistral (7B) and Phi-3-mini (3.8B) perform substantially worse, with maximum pass@1 of 2.6 and 8.0 while they perform comparable as large LLMs for simple mathematical reasoning task~\cite{DBLP:journals/corr/abs-2404-14219}. 
This substantial performance gap highlights the limitations of conventional LLMs in solving complex mathematical problems. 
It demonstrates the efficacy of MC-NEST's advanced sampling strategies, which enhance LLM precision even in ZS and FS settings. 
For the MathOdyssey dataset, GPT-4o shows a reasonable performance with pass@1 from 8.0 to 16.6 with different prompting, but MC-NEST achieves a pass@1 of 13.3. 
In contrast, Mistral (7B) and Phi-3-mini (3.8B) again underperform, with Mistral (7B) struggling notably on FS10 with 0.6.
The results from both datasets support that MC-NEST significantly improves the performance of existing LLMs across different prompting strategies and methods, including ZS, FS3, FS5, FS10, and ZSCoT prompting and MCTSr. Particularly in the MCTSr context, MC-NEST's utilization of probability distribution strategies ensures fairness in the refinement and decision-making processes, facilitating higher pass@1 and robustness in predictions.
By employing probability distribution strategies within its MC-NEST rollouts, MC-NEST effectively mitigates the solution variance observed in traditional prompting approaches with consistent and reliable outcomes.

In \autoref{fig:figure1} and \autoref{tab:domain_difficulty_solved}, we analyze MC-NEST's problem-solving across domains and difficulty levels for both datasets. On AIME, MC-NEST with rollout 4 works well in Algebra (15 medium problems, 6 hard problems), performs moderately in Combinatorics (2 medium problems, 5 hard problems), and effectively addresses Geometry (7 medium problems, 6 hard problems). It achieves its best results in Number Theory (8 medium problems, 9 hard problems), excelling in structured, computation-heavy problems. On MathOdyssey, with rollout 24, performance drops: Algebra (4 medium problems, 3 hard problems), Combinatorics (1 medium problem, 2 hard problems), and Geometry (2 medium problems, 1 hard problem). Number Theory remains relatively stronger (4 medium problems, 3 hard problems), highlighting LLM's self-refine and self-evaluation limitations in less structured, intuition-driven tasks.

\autoref{fig:figure2} compares the Pass@1 performance of various prompting strategies, MCTSr, and MC-NEST on the AIME and MathOdyssey datasets.
MC-NEST solves the highest problem on AIME with 58, significantly outperforming then others.
On MathOdyssey, however, its performance drops to 20, while ZSCoT performs best with 25.
This contrast underscores the differing problem-solving demands of the datasets. Monte Carlo Tree Search-based methods, such as MC-NEST, excel in structured, multi-step reasoning tasks like those in AIME but struggle with the intuition-driven, less formal problems in MathOdyssey.
In contrast, heuristic-driven approaches like ZSCoT prompting leverage pre-trained knowledge and generalization, making them more effective for MathOdyssey.
These results emphasize the importance of customizing search strategies to the inherent structure of mathematical problem distributions.

\begin{table}[htbp]
    \centering
    \small 
    \setlength{\tabcolsep}{4pt} 
    \begin{tabular}{>{\centering\arraybackslash}p{1.8cm}>{\centering\arraybackslash}p{1.4cm}>{\centering\arraybackslash}p{1.4cm}>{\centering\arraybackslash}p{1.4cm}>{\centering\arraybackslash}p{1.2cm}>{\centering\arraybackslash}p{1.4cm}>{\centering\arraybackslash}p{1.4cm}>{\centering\arraybackslash}p{1.4cm}>{\centering\arraybackslash}p{1.2cm}>{\centering\arraybackslash}p{1.4cm}}
        \toprule
        \textbf{Model} & \multicolumn{4}{c}{\textbf{AIME}} & \multicolumn{4}{c}{\textbf{MathOdyssey}} & \textbf{Average} \\
        \cmidrule(lr){2-5} \cmidrule(lr){6-9}
         & \textbf{CS} & \textbf{CY} & \textbf{OY} & \textbf{MR} 
         & \textbf{CS} & \textbf{CY} & \textbf{OY} & \textbf{MR} 
         & \\
        \midrule
        GPT4o & 87.24\% & 81.20\% & 79.48\% & 81.29\% & 89.28\% & 86.42\% & 80.35\% & 86.78\% & \textbf{84.00\%} \\
        Phi-3-mini & 88.33\% & 82.91\% & 77.50\% & 82.91\% & 83.75\% & 82.50\% & 78.75\% & 80.00\% & \textbf{82.08\%} \\
        \bottomrule
    \end{tabular}
    \caption{Evaluation of models based on Completeness (CS), Clarity (CY), Optimality (OY), and Mathematical Rigor (MR) under Human and GPT-4 Evaluation frameworks.}
    \label{tab:tutoreval_scores}
\end{table}

    \subsection{Quality Check}

    \autoref{tab:tutoreval_scores} presents a detailed evaluation of GPT-4o and Phi-3-mini based on Completeness, Clarity, Optimality, and Mathematical Rigor across both of the datasets. GPT-4o demonstrates superior overall performance with an average score of 84.0\%, surpassing Phi-3-mini's 82.08\%. Specifically, GPT-4o excels in Clarity and Mathematical Rigor on the MathOdyssey dataset, achieving scores of 86.42\% and 86.78\%, respectively, highlighting its adeptness in handling complex mathematical reasoning with greater precision. Phi-3-mini, while slightly behind, maintains a solid performance in Completeness and Clarity on the AIME dataset, indicating its capability in structured problem-solving contexts. The marginal differences in Optimality and Mathematical Rigor between the LLMs suggest that if any LLM can solve a problem using MC-NEST, the quality of the solution remains comparable. This underscores the importance of advanced methods like MC-NEST to further enhance these aspects and bridge the gap in solution quality.

    \subsection{Empirical Observations}

    Our proposed method, MC-NEST, demonstrates strong performance on structured mathematical problems such as the AIME dataset, achieving a Pass@1 of 38.6 with GPT-4o outperforming both ZS and ZSCoT. This highlights its effectiveness in tasks requiring iterative refinement, formal reasoning, and optimality search. Similarly, MC-NEST with Phi-3-mini achieves a Pass@1 of 7.33, surpassing ZSCoT's 1.30. However, this approach struggles to generalize to the MathOdyssey dataset, where problems demand comprehensive pattern recognition and intuition-driven reasoning. On MathOdyssey, MC-NEST records a significantly lower Pass@1 of 13.3 with GPT-4o and only 2.0 with Phi-3-mini, while ZSCoT achieves the highest performance with a Pass@1 of 16.6 on GPT-4o. This disparity underscores ZSCoT's strength in leveraging pre-trained heuristic generalization over step-by-step logical expansion. Smaller LLMs, such as Phi-3-mini, further highlight this limitation due to their constrained capacity for iterative refinement, often failing to improve correctness when subjected to critique. This imbalance stems from MC-NEST's reliance on self-refine and Monte Carlo-based backpropagation, which confines it to a limited local search space. In contrast, ZSCoT benefits from the LLM's emergent reasoning capabilities, enabling conceptual leaps. Also, while MC-NEST works well in domains requiring rigorous multi-step deduction, its structured search paradigm struggles to adapt to the less formal, intuition-driven problems characteristic of MathOdyssey.

\section*{Limitations and Future Directions}
\label{sec:Limitations_and_Future_Directions}

\paragraph{Computational Trade-offs} 
MC-NEST introduces a multi-round search and refinement strategy that yields robust performance on complex mathematical reasoning tasks. However, this comes at the cost of increased inference time and computational overhead compared to simpler prompting methods such as ZS and ZSCoT. While this trade-off enables more consistent and structured reasoning, future work should explore optimizations to reduce cost without sacrificing performance. Quantitative analysis of the runtime-performance trade-off would further guide deployment in resource-constrained environments.

\paragraph{Scope of Baseline Comparisons} 
This work focuses on comparisons with standard prompting-based baselines (ZS, FS, ZSCoT) to isolate the contribution of our proposed search strategy. However, the absence of structurally aligned baselines such as Chain-of-Thought (CoT), Program-of-Thought (PoT)~\cite{chen2022program}, and Tree-of-Thought (ToT)~\cite{yao2023tree} limits the scope of comparative analysis. These approaches share similar goals of multi-step and tree-based reasoning and have demonstrated strong performance on similar tasks. Future work will incorporate these advanced baselines to better contextualize MC-NEST's effectiveness and ensure more rigorous benchmarking.

\paragraph{Problem Complexity and Generalization} 
MC-NEST excels on complex, structured benchmarks like AIME, where multi-step logical reasoning is critical. However, it underperforms on simpler or intuition-driven benchmarks such as GSM8K~\cite{cobbe2021training}. This reflects the method's design, which favors iterative symbolic reasoning over heuristic shortcuts. We acknowledge this limitation and plan to explore hybrid techniques that integrate MC-NEST's structured search with strategies better suited to intuitive or high-level abstraction tasks.


\paragraph{Failure Modes and Qualitative Insights} 
MC-NEST struggles on tasks requiring holistic insight or intuition, as seen in its lower performance on MathOdyssey relative to ZSCoT. This indicates a limitation in adapting its refinement-based strategy to non-symbolic reasoning. To better understand these limitations, we plan to augment our analysis with qualitative studies, including: (1) visualizing search patterns, (2) case studies comparing solution trajectories, and (3) error analyses on representative failure cases. These insights will inform future improvements, including hybrid approaches combining structured search with more flexible reasoning mechanisms.

\section*{Conclusion and Future Work}

In this work, we proposed MC-NEST, a novel framework that integrates probabilistic node selection with self-refinement and self-evaluation mechanisms to enhance the mathematical reasoning capabilities of LLMs. By leveraging a Monte Carlo tree search paradigm enriched with structured decision-making, MC-NEST achieves impressive performance on challenging mathematical benchmarks such as AIME. Our results demonstrate that MC-NEST significantly improves the accuracy and consistency of solutions, particularly on complex, multi-step reasoning tasks. However, this performance gain comes with increased computational cost and a dependency on larger LLMs, limiting its effectiveness on smaller LLMs like Mistral (7B). Additionally, MC-NEST shows reduced performance on intuition-based problems, revealing its current specialization in formal, structured reasoning tasks. To address these limitations and broaden the applicability of MC-NEST, we outline the following future directions:
\begin{itemize}
    \item \textbf{Dataset Expansion:} Broaden evaluation on diverse Olympiad-style and real-world mathematical datasets to better assess generalization and robustness.
    \item \textbf{LLMs Diversity:} Evaluate MC-NEST across a wider range of LLM architectures, including emerging LLMs such as OpenAI’s o1~\cite{DBLP:journals/corr/abs-2409-18486} and DeepSeek-R1~\cite{guo2025deepseek}, to study scalability and transferability.
    \item \textbf{Efficiency Optimization:} Develop lightweight variants of MC-NEST and explore pruning or rollout strategies to reduce inference cost without degrading performance.
    \item \textbf{Stronger Baselines:} Incorporate comparisons with structurally aligned reasoning methods such as CoT, PoT, and ToT to better contextualize MC-NEST's contributions.
    \item \textbf{Hybrid Approaches:} Investigate integration with intuition-oriented reasoning frameworks to improve performance on tasks that require abstract or heuristic thinking.
    \item \textbf{Applied Domains:} Extend MC-NEST to real-world domains such as automated theorem proving, symbolic mathematics, and scientific discovery where high-reliability reasoning is essential.
\end{itemize}

Through these efforts, we aim to build upon the foundation established by MC-NEST and contribute to the development of general-purpose reasoning frameworks that combine structured search,  self-refining strategies, and LLM capabilities in a unified and efficient manner.








\section*{RESOURCE AVAILABILITY}


\subsection*{Lead contact}


Requests for further information and resources should be directed to and will be fulfilled by the lead contact, Gollam Rabby (gollam.rabby@l3s.de).




\subsection*{Data and code availability}



The study was carried out exclusively using open-source datasets and software packages. All scripts, outcomes, post-processed datasets, and features will be accessible to the public at \url{https://github.com/corei5/MC_NEST_GPT}.  The dataset utilized for this study is available on Kaggle and Hugging Face. Interested readers and researchers can access the dataset via the following links: 1) AIME Problems (1983--2024) on Kaggle: \url{https://www.kaggle.com/datasets/tourist800/aime-problems-1983-to-2024} and 2) MathOdyssey Dataset on Hugging Face: \url{https://huggingface.co/datasets/MathOdyssey/MathOdyssey}

\section*{ACKNOWLEDGMENTS}


We acknowledge the support of the KISSKI project (funding no. 01IS22093C) for providing computational resources, which will enable us to extend this research in the future.

\section*{AUTHOR CONTRIBUTIONS}



This work was carried out through close collaboration among all authors. G.R. designed and implemented the algorithm, led the experiment, analyzed the results, and wrote the manuscript. F.K. was responsible for implementing and conducting the prompting-based experiments. S.A. played a significant role in conceiving the algorithm design and contributed to the writing of the manuscript.




\section*{DECLARATION OF GENERATIVE AI AND AI-ASSISTED TECHNOLOGIES}


During the preparation of this work, the author(s) used ChatGPT in order to improve grammatical clarity and correct language errors. After using this tool, the author(s) reviewed and edited the content as needed and take(s) full responsibility for the content of the publication.

\bibliography{references}

\bigskip


\newpage

\section{Appendix}








\subsection{Hypothesis}

\paragraph*{\textit{Hypothesis 1:} Probability Distribution-Guided Node Selection Enhances Search Quality}

Incorporating a probability distribution-based strategy in the MCTS node selection process will improve the exploration-exploitation trade-off, achieve higher cumulative reward, and prevent convergence to local optima. Specifically, for any set of child nodes \( \mathcal{N} = \{ N_1, N_2, \dots, N_k \} \), we expect the following:

\begin{itemize}
    \item \textbf{Improved Exploration:} The utility of less-visited nodes \( N_i \) benefits from their interactions with other nodes \( N_j \in \mathcal{N}, j \neq i \), such that node selection maximizes:
    \[
    N^* = \arg \max_{N_i \in \mathcal{N}} \left( \text{Utility}(N_i, \mathcal{N} \setminus \{N_i\}) \right),
    \]
    where \( \text{Utility}(N_i, \cdot) \) captures the contextual value of node \( N_i \) in relation to its peers.

    \item \textbf{Avoidance of Local Optima:} The node selection process discourages premature convergence by maintaining a form of equilibrium:
    \[
    \forall N_i, N_j \in \mathcal{N},\]
    
    \[\quad Q(N_i) + \lambda \cdot \frac{\log(\sum_{j \in \mathcal{N}} \text{visits}(N_j))}{\text{visits}(N_i)} + \alpha \cdot \frac{1}{|\mathcal{N}|} \geq Q(N_j) + \lambda \cdot \frac{\log(\sum_{k \in \mathcal{N}} \text{visits}(N_k))}{\text{visits}(N_j)} + \alpha \cdot \frac{1}{|\mathcal{N}|}
    \]
    ensuring that no single node has a unilateral advantage.

    \item \textbf{Higher Cumulative Reward:} The cumulative reward \( R(t) \) accrued by following the probability-distribution-guided policy is expected to exceed that of standard UCT-based strategies:
    \[
    R(t) = \sum_{i=1}^t Q(N^*(i)) \quad \text{and} \quad R(t) \geq R_{\text{UCT}}(t),
    \]
    where \( N^*(i) \) is the node selected at step \( i \), and \( Q(\cdot) \) denotes the associated reward.

    \item \textbf{Stable Decision Making:} The node selection process accounts for both absolute and relative performance through a normalized distribution:
    \[
    P(N_i \mid \mathcal{N}) = \frac{ \exp \left( \text{Utility}(N_i, \mathcal{N} \setminus \{N_i\}) \right)}{\sum_{j \in \mathcal{N}} \exp \left( \text{Utility}(N_j, \mathcal{N} \setminus \{N_j\}) \right)}.
    \]
\end{itemize}

This hypothesis is tested by comparing MC-NEST with traditional MCTS baselines (e.g., UCT) on exploration efficiency and final solution quality.

\paragraph*{Hypothesis 1 Proof: Probability Distribution in Node Selection}

In the context of the MC-NEST, a key component of the decision-making process during node selection is the adoption of a probability distribution strategy. This section presents the formal derivation and application of the probability distribution strategy used in node selection, demonstrating its role in guiding the search for optimal solutions. The probability distribution strategy within our framework is employed to select the optimal node from the set of child nodes of the root node. The strategy is grounded in game theory, where each node acts as a player, and the goal is to identify a selection where no node has an incentive to unilaterally change its strategy, thereby ensuring a stable and optimal decision-making process.

Let \( \mathcal{N} = \{ N_1, N_2, \dots, N_k \} \) be the set of candidate child nodes of the root node, where each \( N_i \) represents a potential action (or proposed solution) in the search space. Each node \( N_i \) has an associated reward \( Q(N_i) \), which is computed based on its historical performance, and a visitation count \( \text{visits}(N_i) \), representing the frequency with which \( N_i \) has been selected.

\paragraph*{Formal Definition of Probability Distribution in Node Selection}

The selection of the optimal node \( N^* \) follows the probability distribution principle, where the objective is to maximize the collective utility of the nodes while maintaining a balance between exploration and exploitation. Formally, we define the probability distribution strategy for node selection as follows:

\[
N^* = \arg \max_{N_i \in \mathcal{N}} \left( \text{Utility}(N_i, \mathcal{N} \setminus \{N_i\}) \right),
\]
where \( N^* \) represents the node selected according to the probability distribution, and the utility function \( \text{Utility}(N_i, \mathcal{N} \setminus \{N_i\}) \) is a function of the reward and the interaction between node \( N_i \) and the remaining candidate nodes. The utility of each node is computed by combining two key factors:
\begin{itemize}
    \item \textbf{Reward-based Exploration:} This term encourages the selection of nodes with higher rewards, ensuring that nodes with higher historical performance are prioritized.
    \item \textbf{Exploration Term:} This term incentivizes the selection of less-visited nodes, ensuring a balanced exploration of the solution space.
\end{itemize}

The utility function is defined as:

\[
\text{Utility}(N_i, \mathcal{N} \setminus \{N_i\}) = Q(N_i) + \lambda \cdot \frac{\log \left( \sum_{j \in \mathcal{N}} \text{visits}(N_j) \right)}{\text{visits}(N_i)} + \alpha \cdot \frac{1}{|\mathcal{N}|},
\]
where:
\begin{itemize}
    \item \( Q(N_i) \) is the reward of node \( N_i \),
    \item \( \lambda \) is a constant that balances the exploration-exploitation tradeoff,
    \item \( \text{visits}(N_i) \) represents the number of times node \( N_i \) has been visited, promoting exploitation of previously promising nodes,
    \item \( \alpha \) is a weight factor that ensures that all nodes are given fair consideration relative to their performance, reflecting the probability distribution principle.
\end{itemize}

The term \( \frac{1}{|\mathcal{N}|} \) ensures that, in the absence of strong prior information, each node is treated equally in the absence of visitation history. This respects the foundational concept of probability distribution, where no node has an incentive to change its strategy given the strategies of the others.

\paragraph*{Derivation of the Probability Distribution Strategy}

The probability distribution condition ensures that each node \( N_i \) selects its strategy (or action) such that no unilateral deviation from this strategy increases its utility. In the context of node selection, the strategy \( N_i^* \) maximizes its own utility considering the actions of the other nodes.

The expected utility for each node \( N_i \) is the sum of its immediate reward \( Q(N_i) \), the exploration term that accounts for the total number of visits to the nodes in the set \( \mathcal{N} \), and a term that provides fairness to all nodes based on their relative performance. The probability distribution condition is derived by ensuring that each node's strategy maximizes its expected reward, given the strategies of the other nodes.

\[
N_i^* = \arg \max_{N_i} \left( Q(N_i) + \lambda \cdot \frac{\log \left( \sum_{j \in \mathcal{N}} \text{visits}(N_j) \right)}{\text{visits}(N_i)} + \alpha \cdot \frac{1}{|\mathcal{N}|} \right),
\]
which balances the exploration of less-visited nodes and the exploitation of high-reward nodes, ensuring that no node can increase its utility by unilaterally changing its strategy.

\paragraph*{Effectiveness Probability Distribution Strategy in MC-NEST}

The probability distribution strategy significantly enhances the node selection process in MC-NEST by providing a dynamic and adaptive approach to balancing exploration and exploitation. Unlike traditional heuristics such as UCT, which rely on fixed exploration-exploitation formulas, the probability distribution strategy adapts to the developed search space. As more simulations are performed and the tree grows, the strategy ensures that high-reward nodes are increasingly exploited, while also allowing for the exploration of new or less-explored regions of the search space.

The probability distribution approach prevents premature convergence to suboptimal solutions by accounting for the interactions between nodes. This process encourages a more holistic exploration of the solution space, increasing the likelihood of identifying the optimal solution. By ensuring that no node has an incentive to deviate from its strategy, the probability distribution effectively guides the search towards regions of the search space that maximize the collective utility of the nodes, leading to improved overall search performance.

\paragraph*{\textit{Hypothesis 2:} Self-Refine and Self-Evaluation Improve Iterative Reasoning Quality}

We posit that embedding self-refinement and self-evaluation mechanisms within the tree structure improves the agent’s ability to iteratively revise suboptimal reasoning trajectories and adapt based on feedback.

\begin{itemize}
    \item \textbf{Self-Refine:} A node updates its decision based on observed deviation from expected outcome:
    \[
    \text{Adjustment}(N_i) = \beta \cdot \left( \text{Outcome}(N_i) - \text{Action}(N_i) \right),
    \quad \text{Action}'(N_i) = \text{Action}(N_i) + \text{Adjustment}(N_i)
    \]

    \item \textbf{Self-Evaluation:} Nodes adjust their internal utility based on their children’s feedback:
    \[
    P(N_i) = \frac{1}{\text{visits}(N_i)} \sum_{j \in \mathcal{N}_i} Q(N_j) + \lambda \cdot Q(N_i)
    \]

    \item \textbf{Combined Adaptation:} The refined utility of a node is jointly influenced by its self-evaluation and the scarcity of its visits:
    \[
    U(N_i) = P(N_i) + \gamma \cdot \left( 1 - \frac{\text{visits}(N_i)}{V_{\text{max}}} \right)
    \]
\end{itemize}

This hypothesis will be evaluated through ablation studies that isolate the contributions of the self-refine and self-evaluate components to final performance and convergence behavior.

\paragraph*{Hypothesis 2 Proof: Self-Refine and Self-Evaluation in Node Selection}

In this section, we provide a detailed mathematical formulation of the Self-Refine and Self-Evaluation strategies employed within MC-NEST. These strategies are critical for refining the search process, ensuring that each node in the search tree continuously adapts and evaluates its performance in light of its interactions with other nodes, thereby guiding the search toward more optimal solutions.

\paragraph*{Self-Refine in Node Decision Making}

Self-Refine is a process where each node adjusts its decision-making strategy based on the observed outcomes from past actions. Let \( N_i \) denote a node in the tree, with a corresponding action \( \text{Action}(N_i) \) and observed outcome \( \text{Outcome}(N_i) \).

Given that the initial action of a node may not always yield the optimal result, self-refine aims to refine the action by considering the discrepancy between the expected and actual outcomes. The adjustment to the node's action is computed as follows:

\[
\text{Adjustment}(N_i) = \beta \cdot \left( \text{Outcome}(N_i) - \text{Action}(N_i) \right),
\]
where \( \beta \) is a constant that scales the adjustment rate. The refined action \( \text{Action'}(N_i) \) is given by:

\[
\text{Action'}(N_i) = \text{Action}(N_i) + \text{Adjustment}(N_i),
\]

This process ensures that each node refines its decision-making process after each iteration, thereby improving its selection and exploration strategies over time. The updated action aligns the node's decision-making with the observed reward and aims for a closer alignment with the optimal strategy.

\paragraph*{Self-Evaluation in Node Selection}

Self-evaluation is the process by which a node evaluates its performance relative to other nodes in the tree, specifically its children, to determine the quality of its decision-making. Let \( \mathcal{N}_i \) be the set of children of node \( N_i \), where each child \( N_j \) has a reward \( Q(N_j) \) based on its own performance. The performance of \( N_i \) is then evaluated relative to the expected rewards from its children. The self-evaluation function for node \( N_i \) is defined as:

\[
P(N_i) = \frac{1}{\text{visits}(N_i)} \sum_{j \in \mathcal{N}_i} Q(N_j) + \lambda \cdot Q(N_i)
\]

where:
\begin{itemize}
    \item \( Q(N_j) \) is the reward associated with child node \( N_j \),
    \item \( \lambda \) is a weight that balances the contribution of \( N_i \) itself versus the contributions of its children,
    \item \( \text{visits}(N_i) \) is the number of times node \( N_i \) has been visited.
\end{itemize}

The self-evaluation function \( P(N_i) \) allows the node to assess its effectiveness compared to its children and adapt its strategy based on this feedback. If \( P(N_i) \) is lower than expected, the node may adjust its strategy to focus more on promising child nodes in subsequent iterations.

\paragraph*{Combination of Self-Refine and Self-Evaluation}

Both self-refine and self-evaluation are integral to the decision-making process in our approach. After reflecting on its past actions, a node uses self-evaluation to assess how well it has performed compared to other nodes. The combined influence of these processes is captured in the updated action and performance evaluation. Let \( U(N_i) \) represent the updated utility of node \( N_i \), which is a combination of its adjusted action from self-refine and its performance from self-evaluation:

\[
U(N_i) = P(N_i) + \gamma \cdot \left( 1 - \frac{\text{visits}(N_i)}{V_{\text{max}}} \right),
\]
where:
\begin{itemize}
    \item \( \gamma \) is a constant that scales the influence of the number of visits relative to the node’s performance
    \item \( V_{\text{max}} \) is the maximum number of visits among all nodes, ensuring that more frequently visited nodes receive a higher utility score.
\end{itemize}
The updated utility \( U(N_i) \) serves as the key metric for selecting the next node to explore. Nodes with higher utility scores are more likely to be selected, which ensures that the search process increasingly focuses on the most promising areas of the solution space while still allowing for necessary exploration.


Incorporating both self-refine and self-evaluation, the node selection process can be described as follows. After performing self-refine and adjusting the action based on the observed outcome, the node performs self-evaluation to refine its utility. The combined update rule for node \( N_i \) is:
\[
\text{Action'}(N_i) = \text{Action}(N_i) + \beta \cdot \left( \text{Outcome}(N_i) - \text{Action}(N_i) \right)
\]
Also, the updated utility \( U(N_i) \) is:
\[
U(N_i) = \frac{1}{\text{visits}(N_i)} \sum_{j \in \mathcal{N}_i} Q(N_j) + \lambda \cdot Q(N_i) + \gamma \cdot \left( 1 - \frac{\text{visits}(N_i)}{V_{\text{max}}} \right)
\]
This recursive refinement of both the action and utility of each node ensures that the search process continually adapts to the feedback from previous decisions, ultimately improving the overall effectiveness of the MC-NEST.



\subsection{Prompts in Experiment}

\begin{tcolorbox}[colback=blue!5!white, colframe=blue!75!black, title=Prompt for Critique and Refinement]
\label{refinement-prompt}
\textbf{Provide a detailed and constructive critique to improve the answer. Highlight specific areas that need refinement or correction.}

\bigskip
\noindent
\textbf{Instruction:} Refine the answer based on the critique. Your refined answer should be a direct and concise solution to the problem.

\bigskip
\noindent
\textbf{Additional Guidelines:}
\begin{itemize}
    \item Your response should not refer to or discuss the criticisms.
    \item Do not repeat the problem statement.
\end{itemize}

\bigskip
\noindent
\textbf{JSON Response format:}
\begin{verbatim}
{
    "thought": "The thought process behind the answer.",
    "answer": "A float representing the answer to the problem."
}
\end{verbatim}

\end{tcolorbox}

\begin{tcolorbox}[colback=blue!5!white, colframe=blue!75!black, title=Prompt for Reward Limit]
Provide a reward score between -100 and 100 for the answer quality, using very strict standards. Do not give a full score above 95. Make sure the reward score is an integer. Return \emph{ONLY} the score.
\end{tcolorbox}

\begin{tcolorbox}[colback=blue!5!white, colframe=blue!75!black, title=Prompt for field classification]

\bigskip
\noindent
\textbf{JSON format prompt:}
\begin{verbatim}
{
    "system": "The user will provide a problem. Find the general field 
    (such as number theory, geometry, etc) of this math problem. Only return
    the general field. Let's think step by step.",
    "user": "<problem>\n{problem}\n</problem>"
}
\end{verbatim}

\end{tcolorbox}

\begin{tcolorbox}[colback=blue!5!white, colframe=blue!75!black, title=Evaluate the quality]

\bigskip
\noindent
\textbf{JSON format prompt:}
\begin{verbatim}
{
    "system": """ You are a helpful assistant specializing in complex 
    mathematics. Your task is objectively scoring mathematical solutions based 
    on given criteria, ensuring reproducibility and fairness. Provide outputs 
    strictly in the requested JSON format, avoiding unnecessary text.""",

    "user": """ Evaluate the quality of two solutions to a mathematical 
    problem (Human Solution and MC-NEST Solution) based on the following
    criteria. Compare how these solutions differ across the criteria,
    highlighting key strengths and weaknesses for each solution. Return only
    the scores for each criterion with their names and a comparison summary in
    a JSON format.

    ### Evaluation Criteria:

    1. **Completeness (100 points)**: Assess whether all necessary steps are 
    included and logically connected. Comparison of completeness for Human and
    MC-NEST solutions.
    2. **Clarity (100 points)**: Check if the explanation and notation are 
    easy to follow and unambiguous. Comparison of clarity for Human and
    MC-NEST solutions.
    3. **Optimality (100 points)**: Measures the efficiency and elegance of the
    solution. Comparison of optimality for Human and MC-NEST solutions.
    4. **Mathematical Rigor (100 points)**: Verifies if all assumptions, 
    justifications, and definitions are stated explicitly and appropriately. 
    Comparison of mathematical rigor for Human and MC-NEST solutions.

    ### Inputs:
    1. **Human Solution**: "Insert human-generated solution here."
    2. **MC-NEST Solution**: "Insert LLM-generated MC-NEST solution here."

    ### Output Format:
    Return the evaluation scores and comparison summary as a JSON object in
    the following structure:
    {
        "Scores": {
            "MC-NEST Solution": {
                "Completeness": score,
                "Clarity": score,
                "Optimality": score,
                "Mathematical Rigor": score
            }
        }
    }
    Generate the scores and comparisons based on the provided inputs and
    criteria. Use hypothetical scores and comparisons if specific solutions
    are not provided.
    """
}

\end{verbatim}

\end{tcolorbox}


\subsection{Additional Results}

\begin{table*}[h]
    \centering
    \small
    \begin{tabular}{p{2cm}|p{6cm}|p{1cm}|p{1cm}|p{1cm}|p{1.7cm}} 
        \hline
        \textbf{Math domain} & \textbf{Problem} & \textbf{Output} & \textbf{ZS} & \textbf{ZSCoT} & \textbf{MC-NEST} \\
        \hline
        Number \newline Theory & Let $S$ be a list of positive integers not necessarily distinct in which the number $68$ appears. The average (arithmetic mean) of the numbers in $S$ is $56$. However, if $68$ is removed, the average of the remaining numbers drops to $55$. What is the largest number that can appear in $S$? & 649 & 56 & - & 649 \\
        \hline
        Geometry & A machine-shop cutting tool has the shape of a notched circle, as shown. The radius of the circle is $\sqrt{50}$ cm, the length of $AB$ is $6$ cm and that of $BC$ is $2$ cm. The angle $\angle ABC$ is a right angle. Find the square of the distance (in centimeters) from $B$ to the center of the circle. size(150); defaultpen(linewidth(0.6) + fontsize(11)); real r=10; pair O=(0,0), A=r*dir(45), B=(A.x, A.y-r); path P=circle(O,r); pair C=intersectionpoint(B--(B.x+r,B.y),P); // Drawing arc instead of full circle //draw(P); draw(arc(O, r, degrees(A), degrees(C))); draw(C--B--A--B); dot(A); dot(B); dot(C); label("$A$",A,NE); label("$B$",B,S); label("$C$",C,SE); & 26 & - & 50 & 26 \\
        \hline
        Algebra & Let $x$, $y$ and $z$ all exceed $1$ and let $w$ be a positive number such that $\log_x w = 24$, $\log_y w = 40$ and $\log_{xyz} w = 12$. Find $\log_z w$. & 60 & 123 & 60 & 60 \\
        \hline
        Combinatorics & One commercially available ten-button lock may be opened by pressing -- in any order -- the correct five buttons. The sample shown below has $\{1,2,3,6,9\}$ as its combination. Suppose that these locks are redesigned so that sets of as many as nine buttons or as few as one button could serve as combinations. How many additional combinations would this allow? & 770 & 770 & 770 & 770 \\
        \hline
        Others & Let $f(x) = |x - p| + |x - 15| + |x - p - 15|$, where $0 < p < 15$. Determine the minimum value taken by $f(x)$ for $x$ in the interval $p \leq x \leq 15$. & 15 & 15 & 15 & 15 \\
        \hline
    \end{tabular}
    \caption{Comparison of LLM approaches on complex mathematical problem solving across domains. This table showcases a variety of complex mathematical problems, spanning domains such as \textit{Number Theory} and \textit{Combinatorics}, with outputs from standard LLM solutions (Zero-Shot and Zero-Shot Chain-of-Thought) and the specialized MC-NEST algorithm. The "Original Output" column presents baseline/human responses, highlighting the progression from standard to MC-NEST algorithm in achieving accurate solutions.}
    \label{tab:eaxmple_result_Table}
\end{table*}

\begin{table*}[htbp]
\centering
\small 
\begin{tabular}{l|c|c|c|c} 
\hline
\textbf{LLM} & \textbf{Size} & \textbf{Dataset} & \textbf{Prompt/Method} & \textbf{Solved Problem} \\
\hline
\multicolumn{5}{c}{\textbf{Proprietary LLMs}} \\
\hline
GPT-4o & - & AIME & Zero-Shot (ZS) & 40 \\
 & & & Few-Shot (3 Examples, FS3) & 44 \\
 & & & Few-Shot (5 Examples, FS5) & 38 \\
 & & & Few-Shot (10 Examples, FS10) & 38 \\
 & & & Zero-Shot Chain-of-Thought (ZSCoT) & 50 \\
 & & & Monte Carlo Tree Self-refine (MCTSr) & 49 \\
 & & & Monte Carlo Self-Refine Tree (MC-NEST) & 58 \\
\hline
Mistral & 7B & AIME & Zero-Shot (ZS) & 2 \\
 & & & Few-Shot (3 Examples, FS3) & 4 \\
 & & & Few-Shot (5 Examples, FS5) & 4 \\
 & & & Few-Shot (10 Examples, FS10) & 4 \\
 & & & Zero-Shot Chain-of-Thought (ZSCoT) & 2 \\
 & & & Monte Carlo Tree Self-refine (MCTSr) & - \\
 & & & Monte Carlo Self-Refine Tree (MC-NEST) & - \\
\hline
Phi-3-mini & 3.8B & AIME & Zero-Shot (ZS) & 7 \\
 & & & Few-Shot (3 Examples, FS3) & 12 \\
 & & & Few-Shot (5 Examples, FS5) & 8 \\
 & & & Few-Shot (10 Examples, FS10) & 10 \\
 & & & Zero-Shot Chain-of-Thought (ZSCoT) & 2 \\
 & & & Monte Carlo Tree Self-refine (MCTSr) & 11 \\
 & & & Monte Carlo Self-Refine Tree (MC-NEST) & 11 \\
\hline
\hline
GPT-4o & - & MathOdyssey & Zero-Shot (ZS) & 21 \\
 & & & Few-Shot (3 Examples, FS3) &  16 \\ 
 & & & Few-Shot (5 Examples, FS5) & 18 \\ 
 & & & Few-Shot (10 Examples, FS10) & 12 \\ 
 & & & Zero-Shot Chain-of-Thought (ZSCoT) & 25 \\ 
 & & & Monte Carlo Tree Self-refine (MCTSr) &  16\\
 & & & Monte Carlo Self-Refine Tree (MC-NEST) &  20\\
\hline
Mistral & 7B & MathOdyssey & Zero-Shot (ZS) & 27 \\
 & & & Few-Shot (3 Examples, FS3) & 28 \\
 & & & Few-Shot (5 Examples, FS5) & 28 \\
 & & & Few-Shot (10 Examples, FS10) & 1 \\
 & & & Zero-Shot Chain-of-Thought (ZSCoT) & 20\\
 & & & Monte Carlo Tree Self-refine (MCTSr) & - \\
 & & & Monte Carlo Self-Refine Tree (MC-NEST) & -\\
\hline
Phi-3-mini & 3.8B & MathOdyssey & Zero-Shot (ZS) & 12 \\
 & & & Few-Shot (3 Examples, FS3) & 21 \\
 & & & Few-Shot (5 Examples, FS5) &  21 \\
 & & & Few-Shot (10 Examples, FS10) & 7 \\
 & & & Zero-Shot Chain-of-Thought (ZSCoT) & 13 \\
 & & & Monte Carlo Tree Self-refine (MCTSr) & 3 \\
 & & & Monte Carlo Self-Refine Tree (MC-NEST) & 3 \\
\hline
\end{tabular}
\caption{Performance comparison of different LLMs with various prompting approaches. Abbreviations: ZS = Zero-Shot Prompting; FS = Few-Shot Prompting; ZSCoT = Zero-Shot Chain-of-Thought Prompting; MCTSr = Monte Carlo Tree Self-Refine (rollout 4 with Greedy Policy); MC-NEST = Monte Carlo Self-Refine Tree (rollout 16 with Importance Sampling Policy).}
\label{tab:result_com}
\end{table*}

        \begin{figure}[ht]
            \centering
            \includegraphics[height=16cm, width=1.0\textwidth]{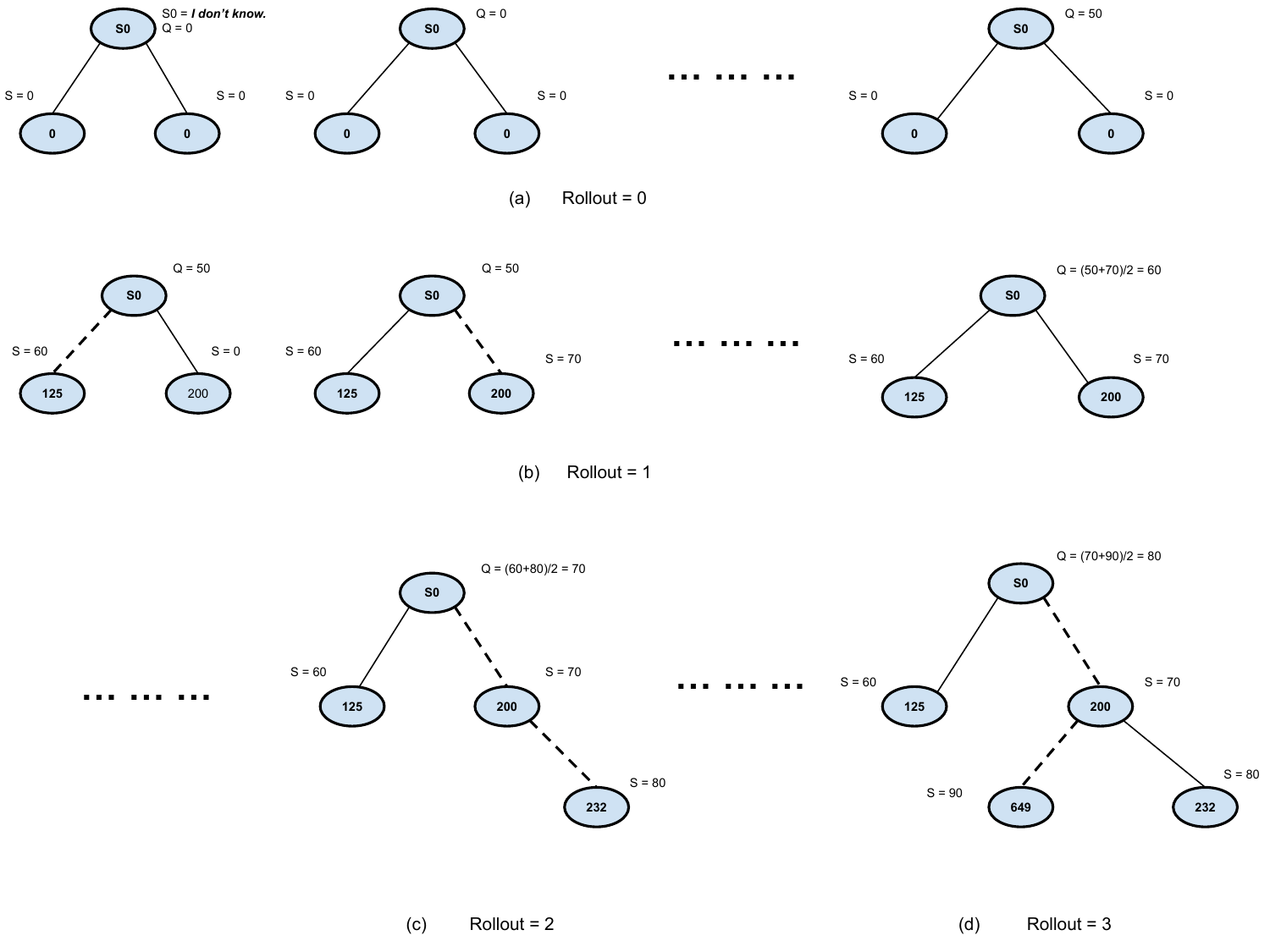}
            \caption{Several MC-NEST processing steps for the problem: Let $S$ be a list of positive integers not necessarily distinct in which the number $68$ appears. The average (arithmetic mean) of the numbers in $S$ is $56$. However, if $68$ is removed, the average of the remaining numbers drops to $55$. What is the largest number that can appear in $S$?}
            \label{fig:MC-NEST_tree}
        \end{figure}

\end{document}